\documentclass[11pt,a4paper]{article}
\usepackage[hyperref]{emnlp2018}
\usepackage{times}
\usepackage{latexsym}
\usepackage{graphicx}

\usepackage{url}

\aclfinalcopy 

\newcommand\blfootnote[1]{%
  \begingroup
  \renewcommand\thefootnote{}\footnote{#1}%
  \addtocounter{footnote}{-1}%
  \endgroup
}

\title{Leveraging Text Repetitions and Denoising Autoencoders in OCR Post-correction}

\author{
  Kai Hakala$^{1,2}$\thanks{\ \ These authors contributed equally.}
  , Aleksi Vesanto$^{1}$\footnotemark[1] , Niko Miekka$^{1}$, \\ 
  \textbf{Tapio Salakoski$^{1}$ and Filip Ginter$^{1}$} \\
  $^{1}$Turku NLP Group, Department of Future Technologies, University of Turku, Finland \\
  $^{2}$University of Turku Graduate School, University of Turku, Finland \\
  \tt \{kahaka,avjves,nikmie,figint,sala\}@utu.fi \\
}

\date{}

\begin{document}
\maketitle
\begin{abstract}
A common approach for improving OCR quality is a post-processing step based on models correcting misdetected characters and tokens. These models are typically trained on aligned pairs of OCR read text and their  manually corrected counterparts. In this paper we show that the requirement of manually corrected training data can be alleviated by estimating the OCR errors from repeating text spans found in large OCR read text corpora and generating synthetic training examples following this error distribution. We use the generated data for training a character-level neural seq2seq model and evaluate the performance of the suggested model on a manually corrected corpus of Finnish newspapers mostly from the 19th century. The results show that a clear improvement over the underlying OCR system as well as previously suggested models utilizing uniformly generated noise can be achieved.
\blfootnote{A Preprint}
\end{abstract}

\section{Introduction}
Analyzing data from scanned documents is a common task in both scientific and industrial applications. This, however is often hindered by the errors produced by the OCR software, which can have major influence on the task at hand. Therefore correcting the errors is a necessity before conducting many of the analyses. OCR improvements are usually separated into three categories: enhancing the OCR software itself, manipulating the scanned images so that the OCR software has an easier job, and lastly correcting the errors afterwards \citep{image-preprocessing,tesseract,postcorrection}. In this paper we focus on the the last of the three. Various different methods have been proposed to achieve this:
\citet{multimodular} use a modular system that combines statistical machine translation with multiple vocabulary level checks to vote for a corrected word. The downside of such systems is the requirement of gold standard training data, i.e. erroneous OCR output and the corresponding manual corrections. 

Recent studies have shown that the need for gold standard data can be circumvented by synthesizing training data by replacing or adding characters with uniform probability and training recurrent neural models to remove the added noise \citep{noisegen}. Older studies on the other hand rely on replacing the given OCR tokens with their most similar hits from a known vocabulary, leading to both misrecognized and other lexical variants being replaced with a unified form \citep{anagram}.

In this paper we build upon both of these ideas: we train a neural sequence-to-sequence (seq2seq) model \citep{seq2seq} able to receive a text span containing OCR errors as an input and producing corrected text as output. To train the model we generate synthetic data by imposing errors on clean text data. Instead of relying on uniformly added noise, we attempt to model the characteristics and systematic nature of the errors produced by a real OCR system. These errors are estimated by aligning similar text spans found from a large OCR read newspaper corpus.

\section{Data}
The main data we attempt to correct is the National Library of Finland's (NLF) \footnote{https://www.kansalliskirjasto.fi/} publicly available collection of approximately three million scanned issues of newspapers and journals ranging from years 1771 to 1910. These scanned documents have been OCR read, but due to many factors, such as old age, bad scans, ink leakage and the hard to read font, Fraktur, the quality is at times very low.
NLF has recently published a corpus of 479 manually corrected issues sampled from the fore-mentioned data.
To compare our noise generation approach to a model trained with gold standard data, we split the data on issue level to training, validation and test sets with 60\% / 20\% / 20\% division.
The training and validation sets are only utilized for a gold standard model, whereas the test set is used as the final evaluation for all compared models. This results in training and test sets of approximately 300K and 110K tokens respectively.

In addition to the NLF data, we also use two clean corpora roughly from the same time period: Finnish literature from the Gutenberg project\footnote{Project Gutenberg. Retrieved February 2018, from www.gutenberg.org.} and the early modern Finnish corpus provided by the Institute for the Languages of Finland (Kotus) \footnote{http://kaino.kotus.fi/korpus/1800/meta/1800\_coll\_rdf.xml}. Although the Kotus corpus aims at covering a large variety of Finnish texts from years 1809 to 1899, we only use the subset concentrating on Finnish newspapers.
Whereas there are plenty of novels in the Gutenberg project data, their style and vocabulary are distinctly different from newspapers. The selected Kotus dataset on the other hand shares both the style and vocabulary of the data we wish to correct, but lacks the size of the Gutenberg corpus. The purpose for selecting these two datasets is to assess how closely related the used corpus has to be to the target text. Even though Gutenberg corpus could benefit from its larger size, in this study we wish to analyze the influence of the used text genre and thus sample 1.5M tokens from both corpora for training.
\begin{figure*}
    \centering
    \small
    \includegraphics[width=2\columnwidth]{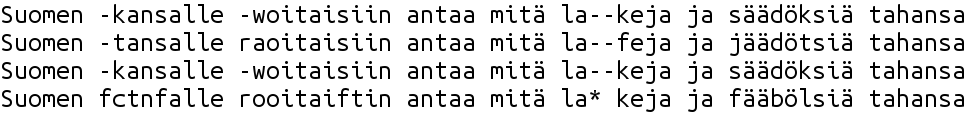}
    \caption{A sample from a single cluster. The most common character for each position is chosen as the representative word and is subsequently aligned against the others to get the replacements. The span can be translated as \emph{"Any laws and regulations could be imposed on Finnish people".}}
    \label{fig:align}
\end{figure*}
\section{Methods}
Our approach to correcting OCR errors is to train a neural seq2seq model that takes in a broken word as input and outputs the corrected word a single character at a time. To this end, we require corrected and aligned OCR texts as training data, which are expensive to manually annotate. To circumvent this, we use clean text and generate noise on top of it. This allows us to generate practically limitless training dataset, given that a clean text corpus is available. In this section we first go through how a realistic noise generation is achieved and then describe how this noise is utilized in training the OCR post-correction model.

\subsection{Noise generation}
We experiment with generating two types of noise for our data. One with uniform random noise, where characters in a word are randomly replaced by an ASCII letter or a number, following the approach by \citet{noisegen}, and a second with noise that attempts to mimic the actual OCR noise in the data. To this end, we calculate a character replacement distribution that tells us how often a given character is replaced by another character. Rather than calculating this from our gold standard documents, we leverage text reuse, a common phenomenon in newspapers \citep{reuse}. We run the OCR data through a text reuse detection method \cite{blast}, originally developed for analyzing how Finnish news spread over time. The tool finds repetition pairs in the text collection and then clusters them. The method is based on sequence alignment algorithms \cite{altschul-blast}, originally developed for gene and protein sequences, and can find text reuse clusters even when there is a significant amount of noise in the data. Each cluster consists of the text spans where the reuse occurs.

We only look at clusters with at least 20 text spans, as the intuition is that if we have enough repetitions of the same piece, we can align them and statistically determine the correct character at each position. Figure \ref{fig:align} shows an example of such alignment. 
For each cluster we first tokenize the spans and group every word in a span with other similar words found in the other spans of the same cluster. Word similarity is measured with Levenshtein distance. We then align all the words in a group with others and calculate the most frequent character for every position. Only words that are common in the reuse cluster are aligned and subsequently used to calculate the distribution. We calculate a replacement occurrence for every character and the distributions are then simply collected by counting the occurrences of such replacements from all clusters.

With the character replacement distribution we generate noise the same way as with the uniform noise. The only difference being that rather than using random ASCII character replacements, we pick a random character from the replacement distribution for the given character. This kind of noise is closer to the actual OCR noise and therefore should make the model a better fit for the task. To distinguish this approach from the uniform noise generation and the manually corrected gold standard data, we refer to it as \emph{realistic} noise in this paper. Examples of the generated noise are shown in Table \ref{fig:noise}.

We also calculate the average character error rate from the distribution and use that as the uniform noise level, i.e. how much noise to generate on top of the clean text. The amount of noise should be set close to the actual noise, or the model's accuracy will suffer \cite{noisegen}. This type of noise rate is not needed for the realistic noise as each character has a probability of being replaced by itself.

\begin{table}[]
\centering
\small
\begin{tabular}{lr}
\hline
\textbf{Type} & \textbf{Text} \\ \hline
GOLD          & kirkonkyl{\"a}n   \\
REAL          & tirkonkvlln   \\
UNI           & kzrkonkbl{\"a}n   \\ \hline
\end{tabular}
\caption{Examples of realistic and uniform noise compared to the Finnish word \emph{kirkonkyl{\"a}n}, genetive form of \emph{"parish"}. }
\label{fig:noise}
\end{table}

\subsection{Denoising Autoencoder}

For correcting the OCR output we have chosen an RNN and attention based seq2seq model as implemented in the Pytorch version of the OpenNMT software \citep{pytorch,opennmt, seq2seq}.
This model uses a stacked LSTM \citep{lstm} encoder to read the input sequence and a similar decoder together with 
a global attention \citep{attention} over the encoder states to form the output predictions. Beam search is used to select the final output sequence from multiple plausible outputs.

The noisified text is fed as an input to the model and the clean version of the corresponding text is used as the desired output. Thus the approach can be seen as a denoising autoencoder model \citep{dae}.
In the case of the gold standard model the original OCR data is used as the input and the manually corrected texts as output.

We use learning rate decay of 0.75 and start the decay only after 25 epochs. Otherwise the default hyperparameters of OpenNMT are used. The training is stopped when the model performance on the validation set is no longer increasing. Note that only the gold standard model uses a manually corrected validation set for this early stopping, whereas all the models based on artificial noise rely on the same noise in the corresponding validation sets. Thus there are no hyperparameters which are optimized for the characteristics of the actual target data. 

\section{Results and Discussion}

We evaluate all the tested models against the manually corrected newspaper texts using character and word error rates (CER/WER) as the measurements. The noise generation approaches suggested in this paper are evaluated against the original OCR output as well as a seq2seq model trained on manually corrected data.
The model trained on gold standard data 
gives us an intuition on how well the selected neural network architecture is able to learn the task to begin with. This is crucial as the model performance is used as a proxy to evaluate the quality of the generated noise.

Four models with generated noise are evaluated in total: one with uniform noise and another with realistic noise for both Gutenberg and Kotus corpora. With both corpora even the uniformly generated noise can be utilized to improve the OCR system as the models trained on Gutenberg and Kotus corpora achieve +3.79pp and +3.6pp increase over the original OCR measured in WER (see Table \ref{tab:results}). This supports the findings of \citet{noisegen}, although the improvements are not as noticeable on our data as in their experiments.

On both used corpora the models trained with realistic noise achieve clearly better performance than the models trained on uniform noise, achieving +5.90pp and +5.69pp in WER over original OCR, for Gutenberg and Kotus respectively. This suggests that our original assumption of uniformly generated noise not being optimal for OCR post-correction is correct and that the OCR produced noise can be estimated efficiently by aligning the repeating text spans found in large text corpora. As both corpora result in fairly similar performance, we conclude that in our current setting selecting a corpus with the same text type and genre as the target data is not critical as long as the texts are roughly from the same time period. 

Although the denoising model based on artificial noise leads to clear improvements in OCR quality, the model trained on manually corrected data achieves far better results with +12.38pp in WER over original OCR. We speculate that this difference is for a large part caused by the fact that in this initial study we have simplified the noise generation to one-to-one character replacements whereas in reality many of the OCR errors are of more complex nature, which can be learned from the manually corrected data. Thus, as a future work, we should look into noise generator models with one-to-many character replacements. As the found text repetitions are aligned on character level by introducing necessary caps, these caps will also reveal the locations where a single character has been replaced by several and our current noise generation model should thus be easily extendable to such error types.

On character level the performance difference between the original OCR and the trained models is more modest, even for the gold standard model. This is mostly due to the fact that the post-processing is focusing on words with only a few incorrect characters.
One possible reason for this is that we are trying to fix independent tokens unaware of their context, a task which is challenging even for humans. The currently used model, however, is not in anyway restricted to single token inputs and we have briefly assessed the plausibility of correcting full sentences at once. The initial observations from this experiment are conflicting as the model trained on gold standard data resulted in abysmal 23.40pp decrease in WER, i.e. the model introduced a plethora of new errors instead of correcting the input. However, a model trained with realistic noise on 2.5M sentences sampled from Gutenberg data resulted in +6.17pp over original OCR, a slight improvement over a model trained with individual tokens. This suggest that a much larger dataset is required for training such model, even if manually corrected data is used. On the other hand this highlights the importance of artificially generated training data, as manually correcting a dataset an order of magnitude larger than the current one would require tremendous resources, whereas artificial data can be generated infinitely by introducing varying noise on top of the same sentences.

Another future work direction would be to look into generative adversarial models \citep{gan}. In this setting there would be no need for aligned texts spans, but instead a generative model could be used to produce noisified texts. An adversarial network would estimate whether the noisified text is artificially generated or a real OCR output. In this approach similar noising and denoising neural models could be utilized, essentially reversing each other, in a cycle-consistent adversarial setting \citep{cyclegan}.

\begin{table}[]
\centering
\small
\begin{tabular}{lrr}
\hline
\textbf{Model} & \textbf{CER} & \textbf{WER} \\ \hline
GUT-UNI        & 0.0943       & 0.2535       \\
GUT-REAL       & 0.0876       & 0.2324       \\
KOT-UNI        & 0.0941       & 0.2554       \\
KOT-REAL       & 0.0902       & 0.2345       \\ \hline
OCR            & 0.0978       & 0.2914       \\
GOLD           & 0.0680       & 0.1676       \\ \hline
\end{tabular}
\caption{Comparison of the tested models and baselines. The performance is measured in character error rate (CER) and word error rate (WER). OCR = FNL OCR without post-processing, GOLD = model trained with gold standard data, GUT = Gutenberg corpus, KOT = Kotus corpus, UNI = uniform noise, REAL = realistic noise.}
\label{tab:results}
\end{table}

\section{Conclusions}

In this study we show that OCR errors can be estimated by leveraging text repetitions found from large OCR read corpora. The quantity and quality of these repetitions is such that a reliable ground truth can be untangled to the extent that measuring the differences between this ground truth and all the corresponding aligned text repetitions can be utilized to approximate the natural error distribution of the underlying OCR system.

We demonstrate that introducing artificial noise on top of clean text following this discovered error distribution can be used as synthetic training data for neural OCR post-correction models. Our study shows that models trained on this type of noise outperform models trained with uniformly generated noise.

The paper also mentions initial results from further experiments with context-aware models and suggests various future work directions.

\section*{Acknowledgments}
This work was supported by Academy of Finland. Computational resources were provided by CSC - IT Center For Science Ltd., Espoo, Finland.

\bibliography{emnlp2018}
\bibliographystyle{acl_natbib_nourl}

\appendix

\end{document}